%% file: main.tex
\documentclass[runningheads]{llncs}

\usepackage{accv}
\usepackage[width=122mm,left=12mm,paperwidth=146mm,height=193mm,top=12mm,paperheight=217mm]{geometry}

\usepackage{accvabbrv}
\usepackage[numbers,sort&compress]{natbib} %

\usepackage{graphicx}
\usepackage{booktabs}

\usepackage[accsupp]{axessibility}  %

\newcommand\samethanks[1][\value{footnote}]{\footnotemark[#1]}

\usepackage{hyperref}

\usepackage{orcidlink}

\begin{document}

\title{TCG-AR: Real-Time Multi-View Augmented Reality for Trading Card Game Streaming} 

\titlerunning{TCG-AR}

\author{
Anthony Cioppa\inst{}\thanks{Equal contributions. Code: \url{https://github.com/ULiege-VIULab/tcg-ar}}\orcidlink{0000-0002-5314-9015}
\and
Antoine Verdonck\samethanks\orcidlink{0009-0006-0590-2275}
\and
Maxim Henry\orcidlink{0009-0007-8899-8723} 
\and \\
Marc Van Droogenbroeck\orcidlink{0000-0001-6260-6487}
\and
Rapha{\"e}l La Rocca\orcidlink{0000-0003-3388-3041} 
}

\authorrunning{A. Cioppa}

\institute{University of Liège, Belgium 
\email{anthony.cioppa@uliege.be}\\
}

\maketitle

\input{sections/0_abstract}

\input{sections/1_introduction}

\input{sections/2_related_work}

\input{sections/3_methodology}

\input{sections/4_dataset}

\input{sections/5_experiments}

\input{sections/6_conclusion}

\bibliographystyle{splncs04-MVD}
\bibliography{bib/abbreviation-short,bib/all}

\include{sections/7_appendix}

\end{document}

%% file: sections/0_abstract.tex
\begin{abstract}
Trading card games are increasingly played and broadcast online, yet live streams remain mostly limited to flat top-down footage of the playing area.
Augmenting such streams with virtual models of the played cards would improve the viewing experience, but most existing systems rely on instrumented playing surfaces and embedded chips, which are costly and impractical for casual players and large-scale events.
In this work, we present \mbox{TCG-AR}, a novel real-time pipeline that augments trading card games using ordinary RGB cameras alone, without any physical markers or specialized hardware.
Our pipeline detects, orients, and identifies the cards on the board, renders virtual content onto each card across all views, and can additionally compose a broadcast-style view that summarizes the game state for spectators, streaming the augmented feeds to standard broadcasting software such as OBS.
To train the detection, orientation, and identification models without manual labeling, we introduce an automatic procedure that generates annotated synthetic training data from a reference set of card images.
Then, we evaluate several trained models on a new manually annotated dataset with real images, analyzing performance and runtime throughput that determine real-world usability.
Overall, by relying only on commodity cameras and hardware, and by open-sourcing all code, models, and datasets, this work aims to serve as a reference for real-time trading card recognition and to make real-time augmented-reality streaming accessible to the broader community of players and streamers.

\keywords{Augmented reality \and Object detection \and Metric learning \and Real-time systems \and Trading card games}
\end{abstract}

%% file: sections/1_introduction.tex
\section{Introduction}
\label{sec:introduction}

\begin{figure}[t]
  \centering
  \includegraphics[width=\linewidth]{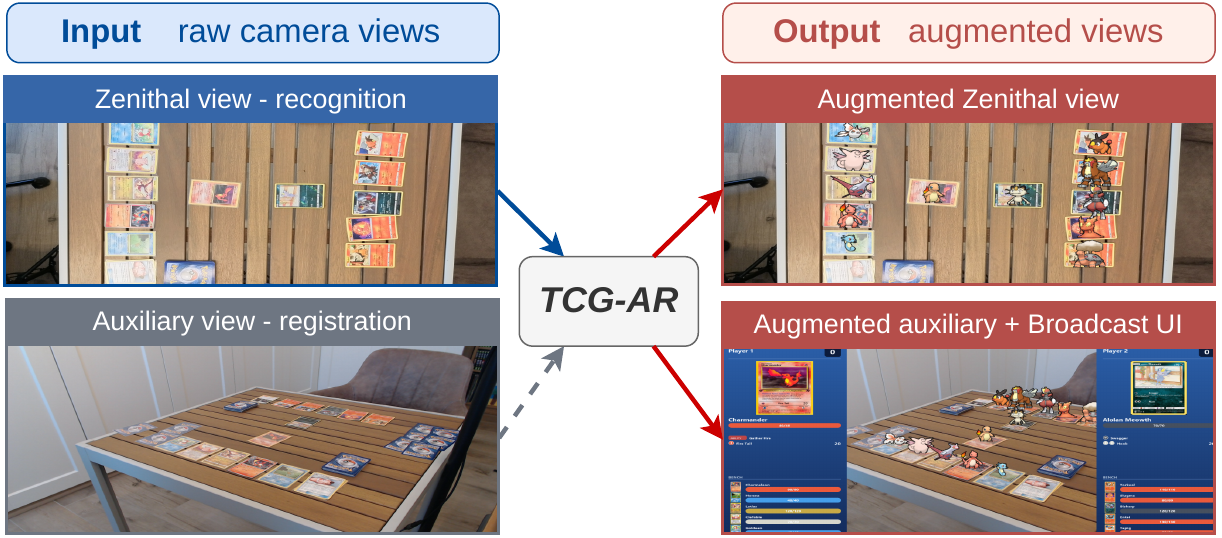}
  \caption{\textbf{Trading Card Game -- Augmented Reality (\mbox{TCG-AR})}. Given one or more camera views of a trading card game (left), \mbox{TCG-AR} detects, orients, and identifies the played cards, and renders a virtual model of the corresponding creature onto each card, in every view and in real time (right). The augmented feeds can be streamed directly to standard broadcasting software. No markers, chips or instrumented surfaces are required. Illustrated here on the Pok\'emon trading card game.}
  \label{fig:teaser}
\end{figure}

Trading card games, such as the Pok\'emon, Yu-Gi-Oh!, Lorcana, One Piece, and Magic: The Gathering are played by millions of people worldwide and have developed a large online presence~\cite{Bigorbit2023ABrief, Statista2025Pokemon}.
Their cultural reach continues to grow, with 2026 marking the thirtieth anniversary of both the Pok\'emon franchise and its trading card game, drawing renewed public attention worldwide~\cite{PokemonCompany2026Anniversary}.
In the current social-media era, players increasingly record and broadcast their games on platforms such as YouTube or Twitch, both for casual sharing and for competitive events~\cite{PokemonChampionships2025Worlds}.
However, these broadcasts remain visually limited: a stream typically shows a flat top-down view of the playing area, in which the cards appear as static pieces of printed cardboard, sometimes supplemented by additional views of the players.
For viewers who are not already invested in the game, this presentation conveys little of the action and offers limited spectacle.%

A natural way to enrich such broadcasts is through Augmented Reality (AR), which integrates virtual content into a view of the real world~\cite{Azuma1997ASurvey, Billinghurst2015ASurvey}.
Applied to a trading card game, AR can replace each played card on the stream with an animated model of the creature it represents, drawing on a long-standing direction in AR gaming that overlays virtual content onto physical play~\cite{Thomas2012ASurvey} to bring the broadcast closer to the experience depicted in the franchise's associated media.
A popular illustration of this idea was demonstrated by the content creator \emph{SuperZouloux}~\cite{SuperZouloux2022LeReve}, whose system displays three-dimensional monster models on top of Yu-Gi-Oh! cards during a live game.
While the result is visually convincing, it is achieved through dedicated hardware: a chip is embedded in every card, and the playing surface is instrumented with readers that detect which card is present at each position.
Such a setup is costly and laborious, as it requires chipping all cards of both players, building a specialized board, and configuring the system, which places it out of reach of casual players and impractical for events that broadcast many games in parallel.
Other systems that augment card games with virtual models also rely on instrumented surfaces~\cite{Lam2006ART} or markers attached to the cards~\cite{Lee2005TARBoard} to locate and recognize them.
More recently, consumer tools have used a webcam or phone camera to identify a single trading card for cataloging~\cite{Eyevo2025Pokemon} or broadcast overlays~\cite{HitScanTCG2025AIPowered,HypeOverlay2026OBSNative}, but these target collection and pack-opening use cases where a card is presented alone and unobstructed, rather than during an actual game where cards are laid out together on a board.

In this work, we address this gap by proposing \mbox{TCG-AR}, a novel real-time pipeline that augments trading card games using ordinary RGB cameras alone, without any physical markers or specialized hardware.
Our pipeline, illustrated in~\cref{fig:teaser}, ingests one or more camera feeds and creates a game state by detecting, orienting, and identifying the cards from a zenithal view. 
The predicted card positions are then mapped into each additional view by registration, allowing a virtual model to be rendered onto every card across all feeds.
From the game state, our pipeline also composes a broadcast-style view that summarizes the match for spectators.
The augmented feeds are then streamed to standard broadcasting software such as OBS, integrating directly into the tools content-creators already use.
To keep the stream fluid despite the computational cost of recognizing the game state, our pipeline separates a fast rendering path, which produces the augmented videos at a high frame rate, from the slower recognition path, which periodically updates the game state.
We design the system to be agnostic to specific trading card games, using the Pok\'emon trading card game only as our demonstration case. 
To do so, we train the detection, orientation, and identification models on synthetic data produced by our automatic generation procedure, avoiding any manual annotation, and evaluate the performance on a collected and manually annotated dataset of real card games.
At run time, our game state recognition path reliably detects, orients, and identifies each card, while the rendering path runs at a high frame rate to place an animated model on every card and stream the augmented feeds to standard broadcasting software over Real Time Streaming Protocol (RTSP).
Beyond its technical design, our aim is also practical: by relying only on commodity cameras and consumer-grade gaming hardware, we seek to make real-time AR streaming accessible to the broader community of trading card game players and streamers.

\noindent\textbf{Contributions.} We summarize our contributions as follows: (i) We present \mbox{TCG-AR}, a real-time pipeline that detects and identifies trading cards from an RGB camera during real gameplay and renders augmented-reality content onto the corresponding cards across one or more views, without any physical markers or instrumented playing surface. \mbox{TCG-AR} further composes a broadcast-style view of the game state and streams the operator-selected feeds to standard broadcasting software via RTSP.
(ii) We propose an automatic, game-agnostic procedure for generating synthetic training data from a reference set of card scans, producing detection, orientation, and identification datasets with annotations obtained at no manual labeling cost, which allows direct adaptation of \mbox{TCG-AR} to any trading card game.
(iii) We collect and manually annotate a real evaluation dataset of the Pok\'emon trading card game, captured under varied conditions, to assess the game state recognition Sim2Real gap.
(iv) We report a quantitative and qualitative study of our pipeline, analyzing detection and identification performance, comparing several model variants for detection, orientation, and identification, and characterizing the streaming runtime throughput.

%% file: sections/2_related_work.tex
\section{Related Works}
\label{sec:related_works}

\subsection{Augmented reality}
Augmented reality aims to embed virtual content within a view of the real world, in real time and spatially registered~\cite{Azuma1997ASurvey}, nowadays extended to modern tracking, display, and interaction technologies~\cite{Billinghurst2015ASurvey}.
Placing virtual content correctly in a moving-camera setting requires continuously estimating the camera pose relative to the scene, the problem at the core of vision-based AR registration~\cite{Marchand2016Pose}.
Classical solutions track the camera geometrically from natural image features~\cite{MurArtal2015ORBSLAM}, while learning-based methods regress the camera pose directly from a single RGB image~\cite{Kendall2015PoseNet,Kendall2016Modelling}.
When the scene is planar and the cameras are fixed, placing overlays across views reduces to estimating a homography between them, classically by detecting and matching keypoints and fitting the transformation robustly with RANSAC~\cite{Fischler1981Random,Chum2005TheGeometric}. 
In this case, hand-crafted features such as SIFT~\cite{Lowe2004Distinctive} and ORB~\cite{Rublee2011ORB} remain widely used, while learned detectors and matchers, including SuperPoint~\cite{DeTone2018SuperPoint}, SuperGlue~\cite{Sarlin2020SuperGlue},  LightGlue~\cite{Lindenberger2023LightGlue}, and LoFTR~\cite{Sun2021LoFTR}, improve matching under low texture and large viewpoint change.
Tabletop and game scenarios have been a recurring target for AR. For example, Rematas~\etal~\cite{Rematas2018Soccer} reconstruct a soccer game from ordinary video to replay it on a tabletop.
For trading card games, early systems rely on markers or sensor-equipped surfaces.
ARTable projects content onto an instrumented table~\cite{Lam2006ART}, and popular hobbyist systems achieve convincing overlays by embedding chips in the cards and reading them with dedicated hardware~\cite{SuperZouloux2022LeReve}.
More recent consumer tools instead recognize cards directly from RGB images, identifying a single card held up to the camera to catalogue it~\cite{Eyevo2025Pokemon} or to overlay its name, image, and market price on a stream~\cite{HitScanTCG2025AIPowered,HypeOverlay2026OBSNative}; these operate on one isolated card at a time and neither localize multiple cards on a board nor register the overlay across viewpoints.
Unlike these, our TCG-AR pipeline recognizes the whole board during play, including overlaid and obstructed cards, from RGB images alone.

\paragraph{Object detection.}
Locating objects in an image is nowadays dominated by deep learning detectors, split between two-stage methods that propose and then classify regions of interest, from R-CNN~\cite{Girshick2014Rich} to Faster R-CNN~\cite{Ren2017Faster}, and one-stage methods that predict boxes directly, such as YOLO~\cite{Redmon2016YOLO}, SSD~\cite{Liu2016SSD}, and RetinaNet with its focal loss~\cite{Lin2017Focal}, alongside efficiency-oriented~\cite{Tan2020EfficientDet} and transformer-based set-prediction designs~\cite{Carion2020EndtoEnd}.
When objects are not axis-aligned, axis-aligned boxes discard their orientation, motivating oriented object detection, studied for aerial and text imagery through rotated-anchor and refinement methods~\cite{Yang2021R3Det,Han2021ReDet}, transformer-based oriented detectors~\cite{Ma2021Oriented-arxiv}, and the Oriented R-CNN detector we adopt, which generates oriented proposals efficiently~\cite{Xie2021Oriented}; a compact classifier on an efficient backbone~\cite{Tan2019EfficientNet} that resolves the residual $180^\circ$ orientation ambiguity.
Training detectors without manual annotation is commonly achieved with synthetic data: domain randomization varies lighting, texture, and pose so that models trained on non-photorealistic renders transfer to real images~\cite{Tobin2017Domain,Tremblay2018Training}, cut-and-paste synthesis composites object crops onto diverse backgrounds with automatically derived labels~\cite{Dwibedi2017Cut,Georgakis2017Synthesizing}, and large rendered datasets serve driving and scene understanding~\cite{Richter2016Playing,Ros2016TheSynthia,Fonder2019MidAir}.
Applied to cards, these ideas have produced convolutional detectors for cluttered poker scenes~\cite{Chen2020Poker}, card classifiers~\cite{Mittal2024Playing}, and automatically generated card datasets for bridge broadcasting~\cite{Wzorek2021Training-arxiv}.
In this work, we detect cards with an oriented detector trained on synthetic data, and evaluate it on real footage. 
 
\paragraph{Identification and re-identification.}
Distinguishing instances within a single object category, and recognizing the same instance again across frames or cameras, is a fine-grained problem that is typically cast as metric learning rather than flat classification. 
Images are mapped to an embedding in which a nearest-neighbor search retrieves the closest reference.
This embedding-and-retrieval paradigm was popularized by Siamese~\cite{Bromley1993Signature, Yi2014Deep} and contrastive~\cite{Hadsell2006Dimensionality} formulations. 
Such paradigms rely on losses such as the triplet, which produces a compact descriptor compared by Euclidean distance~\cite{Schroff2015FaceNet}, and was subsequently refined through better losses and sampling of informative pairs and triplets~\cite{Sohn2016Improved,Wu2017Sampling,Wang2019MultiSimilarity}, proxy- and classification-based objectives~\cite{MovshovitzAttias2017NoFuss,Deng2022ArcFace,Zhai2018Classification-arxiv}, and surveyed comparisons~\cite{Musgrave2020AMetric}.
The same formulation underlies object re-identification, where an instance must be matched across views despite changes in appearance~\cite{He2021TransReID,Ye2024Transformer,Somers2025Person, Somers2023Body}, a setting well suited to many classes with few samples each.
A trading card game is an extreme such case: tens of thousands of visually similar cards, with new ones released over time, which a fixed-output classifier cannot accommodate.
For example, the Pok\'emon trading card game currently comprises up to $20{,}360$ distinct cards, orders of magnitude more than a standard deck of playing cards ($52$).
In this work, we therefore identify cards by retrieval against per-card reference embeddings, so the label set can grow without retraining, and reuse the same descriptors to re-identify cards in the stream.

%% file: sections/3_methodology.tex
\section{Methodology}
\label{sec:methodology}

\subsection{Pipeline overview and design constraints}
\label{sec:overview}

\begin{figure}[t]
    \centering
    \includegraphics[width=\linewidth]{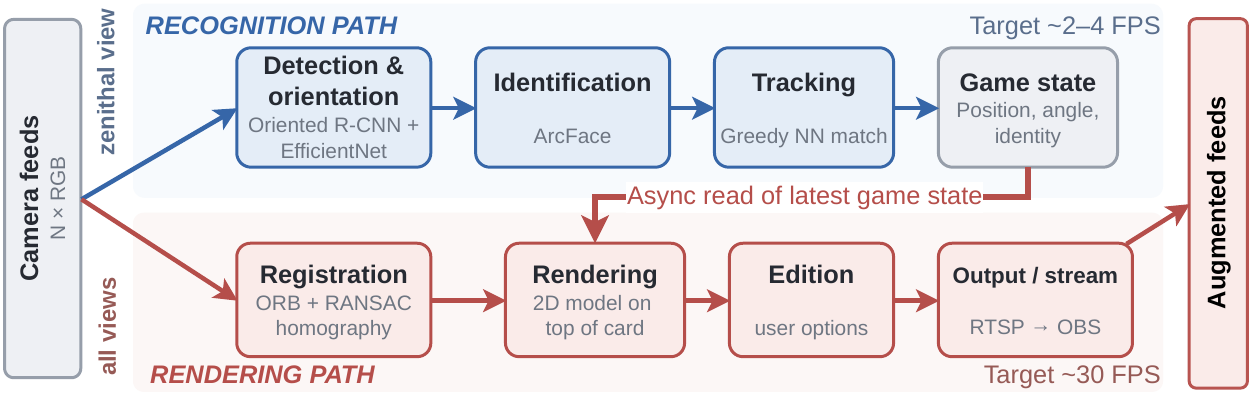}
    \caption{\textbf{Overview of our \mbox{TCG-AR} pipeline.} \mbox{TCG-AR} runs two concurrent paths. The game state recognition path (top) operates on a zenithal view alone: it detects and orients the cards, then identifies and tracks them to estimate the game state, \ie, the set of cards on the board with their identity, position, and orientation. The rendering path (bottom) operates on all views: it registers the auxiliary views to the zenithal reference, renders a virtual model onto each card from the current game state, applies user-selected edition options, and streams the augmented feeds to standard broadcasting software. The rendering path reads the latest game state asynchronously, so the displayed video stays fluid even when recognition is slower.}
    \label{fig:pipeline}
\end{figure}

\mbox{TCG-AR} takes as input one or more RGB camera feeds of a trading card game played on a flat surface, and produces the augmented feeds with a virtual model rendered on top of each played card.
We formulate the task as game state recognition, defined as the set of cards currently on the board together with their position, orientation (\cref{sec:detection}), and identity (\cref{sec:identification}).
Game state recognition is performed on a single zenithal view, where the cards appear with a near-frontal geometry.
The remaining cameras serve only for output: they display the augmented result from additional viewpoints.
The card positions estimated in the zenithal view are mapped into each auxiliary view through image registration (\cref{sec:registration}), where the virtual models are then drawn (\cref{sec:rendering}). 

Our \mbox{TCG-AR} pipeline, illustrated in \cref{fig:pipeline}, runs two concurrent paths so that the displayed frame rate is not tied to the cost of game state recognition.
The recognition path updates the game state by performing detection, orientation, and identification of the cards on the zenithal view.
The rendering path acquires frames, renders the current game state onto them, and streams the result, reading the most recent state asynchronously so the video stays fluid even when recognition is slower.
The two paths face different timing requirements.
The rendering path drives what the viewer sees and targets the $30$ frames per second common to streaming setups.
The recognition path is bound only by latency, since the board typically changes every several seconds: the output must reflect a change quickly enough to go unnoticed.
Using the average human Simple Reaction Time (SRT) of roughly $250$\,ms as a reference~\cite{Woods2015Factors}, we target a game state refresh of about $4$ frames per second as comfortable.
This budget leads the recognition path to favor models and matching strategies that run on commodity hardware over heavier, more accurate alternatives; we motivate these choices per stage and report the resulting frame rates in \cref{sec:experiments}.
While our figures and experiments use the Pok\'emon trading card game, \mbox{TCG-AR} assumes only a reference set of card images, and applies to other trading card games without modification.

\subsection{Card detection and orientation}
\label{sec:detection}

The first stage of the game state recognition path aims to locate the cards in the zenithal view.
While cards are rectangular and rigid, making them well suited to detection, they may not always be axis-aligned: players place and rotate cards freely, and in several trading card games, a rotation may encode a change in a card's status.
An axis-aligned bounding box would therefore discard the orientation of each card, whereas the subsequent identification stage (\cref{sec:identification}) needs an upright, canonical crop to compare a card against the reference set.
We consequently detect cards with an oriented object detector, which outputs for each card a rotated bounding box parameterized by its center, width, height, and in-plane angle.
We compared several detectors and chose the Oriented R-CNN~\cite{Xie2021Oriented} model for this stage, a two-stage detector that generates oriented region proposals and refines them into rotated boxes.
We adopt a ResNet-50~\cite{He2016DeepResidual} backbone with a feature pyramid~\cite{Lin2017Feature}, rather than a heavier backbone or a transformer-based detector, because this configuration meets the timing budget while retaining sufficient performance (see \cref{sec:experiments}).
The detector is trained for a single object class, ``card'', since this stage only needs to separate cards from the background and the playing surface.

However, while the detector localizes each card, it only resolves its orientation up to a sign: the rotated box fixes the card's axis but not which of its two ends points up, leaving a $180^\circ$ ambiguity.
This ambiguity matters because %
the identification stage expects an upright one.
We resolve this orientation task with a lightweight orientation classifier applied to each detected card crop, which predicts whether the crop is upright or rotated by $180^\circ$. 
Crops predicted as inverted are then flipped before identification.
We investigated several architectures and chose the EfficientNet-B0 network~\cite{Tan2019EfficientNet} whose classification head is replaced by a two-way output.
Together, the two components turn the zenithal frame into a set of upright, canonically oriented card crops with known positions, which the identification stage can use to determine each card's identity.

\subsection{Card identification}
\label{sec:identification}

Given the upright card crops produced by the detection and orientation stage, the identification stage assigns the identity of the card it depicts to each crop, drawn from the reference set of all cards in the game.
Trading card games may contain on the order of tens of thousands of distinct cards, many of which differ only in small regions such as a name, a numeric value, or an illustration.
Furthermore, the set of cards grows over time as new ones are released every $3$ to $6$ months.
Hence, training a flat classifier over so many near-identical classes, and retraining it whenever the set changes, is impractical.
We therefore cast identification as a metric learning task, following the embedding-and-retrieval approach popularized for face recognition~\cite{Schroff2015FaceNet}.
A model maps first each card crop to a fixed-length descriptor, and then a card is identified by comparing its descriptor to a set of reference descriptors, one per known card, computed once from the reference card images.
Identification reduces to a nearest-neighbor search in descriptor space, so adding or removing cards only requires inserting or deleting reference descriptors, with no retraining.
We use a ResNet-50~\cite{He2016DeepResidual} backbone, from which the descriptor is produced by one of two heads that we compare empirically.
The first head follows the original triplet formulation~\cite{Schroff2015FaceNet}: a fully connected layer maps the backbone features to a $128$-dimensional descriptor trained with a triplet margin loss. 
Each triplet consists of a reference card image as anchor, a second view of the same card as positive, and a different card as negative, and the loss drives same-card descriptors closer together than different-card descriptors by at least a fixed margin under the Euclidean distance.
The second head, which we adopt by default, replaces the triplet objective with the additive angular margin loss of ArcFace~\cite{Deng2022ArcFace}: each card defines a class whose $512$-dimensional embedding is constrained to a hypersphere, an angular margin is enforced between classes during training, and cards are matched at test time by cosine similarity. 
This optimizes the embedding against all cards jointly rather than from sampled triplets, which we find yields a more robust descriptor on real crops. 
Both heads share the same backbone, the same one-per-card reference descriptors, and the same nearest-neighbor retrieval to ensure fair comparison.
Since the ArcFace head defines one class per card and needs only the reference images, whereas selecting difficult negatives for the triplet head~\cite{Hadsell2006Dimensionality} uses game-specific metadata (detailed for Pokémon in the supplementary material), deploying ArcFace keeps both the data generation and the identification model agnostic to the specific trading card game.

Comparing each crop against the descriptors of all cards in the game is a source of avoidable errors, since most of those cards cannot appear in a given match.
A game is typically played from decks that together contain only a small fraction of the existing cards.
We can exploit this by restricting the nearest-neighbor search to the cards declared in the players' deck lists, which reduces the candidate set from the full reference set to a few tens/hundreds of cards.
This restriction removes the possibility of matching to a card that is not in play, which improves identification accuracy.
Finally, since identification is run per frame, but the cards on the board persist across frames, recomputing every identity at every recognition step is wasteful and prone to flicker.
Therefore, we associate the cards detected in the current frame with those in the previous game state by a greedy nearest-position matching to improve identification stability.%

\subsection{Multi-view registration}
\label{sec:registration}

To render a model onto a card in an auxiliary view, the card's location in that view must be known.
Rather than running detection and identification separately in every view%
, we recover the card locations in the auxiliary views by mapping the zenithal positions into them.
Since cards lie on a common planar surface and the cameras are static, each auxiliary view is related to the zenithal view by a homography, \ie, a $3\times3$ projective transformation between the two image planes.
We estimate each homography once, at initialization, from a pair of frames showing the same board.
The automatic procedure first detects keypoints in both frames with ORB~\cite{Rublee2011ORB}. %
From correspondences, the homography is fitted with RANSAC~\cite{Fischler1981Random}.
When the automatic procedure fails to find enough correspondences, for instance under repetitive or low-texture backgrounds, we fall back to a manual step in which the user provides the corresponding pairs.

\subsection{Rendering, edition, and streaming output}
\label{sec:rendering}

The rendering path turns the current game state into the augmented video.
For each frame, and for each card present in the game state, it draws a virtual model of the entity the card represents on top of each card. %
Virtual content is drawn from a model database that pairs each card identity with a renderable asset.
We render two-dimensional assets, compositing a flat, animated sprite onto the card region as well as broadcast-style information about the game, such as the current active card, the bench, as well as card information (\eg, type, health, attacks, etc.). 
Further details are provided in supplementary material.
Finally, the output stage makes the augmented feeds available to external software.
Each enabled feed is emitted as a Real-Time Streaming Protocol (RTSP) stream, in up to three forms per camera, namely its raw image, its augmented version, and its broadcast composition, with only the operator-selected forms produced and streamed. 
Rather than provide editing and broadcasting tools of its own, the system delegates these to existing streaming software: the RTSP streams can be ingested directly by common tools such as OBS, where the operator composes the final broadcast.
This keeps \mbox{TCG-AR} focused on producing the augmented views and lets it fit into the workflows players and streamers already use.

%% file: sections/4_dataset.tex
\section{Datasets}
\label{sec:datasets}

\begin{figure}[t]
  \centering
  \includegraphics[width=\linewidth]{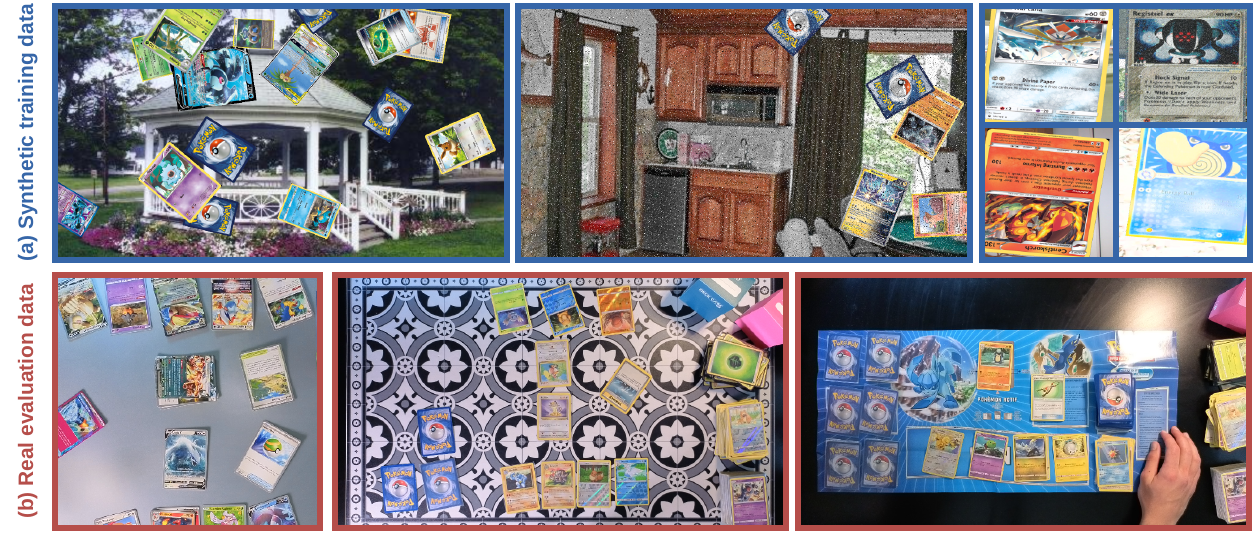}
  \caption{\textbf{Examples from our datasets.} (a) Synthetic training data generated automatically from the reference card images: a detection sample with multiple cards composited onto a textured background, single-card crops for orientation, and identification crops of one card under varied lighting, color, and noise. (b) Real evaluation scenes captured under different lighting and board conditions, manually annotated with oriented boxes and card identities. Illustrated for the Pok\'emon trading card game.}
  \label{fig:dataset-examples}
\end{figure}

The detection, orientation, and identification models are trained on synthetic data that we generate automatically, with no manual annotation.
Our generation procedure takes as its only game-specific input a reference set of card images, one canonical image per card, which is readily available for trading card games~\cite{Scrydex2025Scrydex,Backes2021PokemonTCGSDK}.
From this reference set, our procedure generates a separate datasets for each task, with the annotations derived directly from how each image is constructed.
We instantiate it on the Pok\'emon trading card game, whose reference set contains $K=20{,}360$ different cards at the time of writing, spanning $1{,}025$ distinct creatures and many more trainer, energy, and object cards.
Since nothing in the procedure is specific to a particular game beyond the reference images, the same generator applies to other trading card games.
We additionally collect a real, manually annotated dataset, used only for evaluation, to measure how well the resulting models transfer to genuine play.
\Cref{fig:dataset-examples} shows examples of both synthetic and real images, and \cref{tab:datasets} summarizes their composition.

\input{tables/datasets}

\subsection{Synthetic data generation}
\label{sec:synthetic}

The detection dataset is built by compositing cards onto background images.
We draw a background photograph from a collection of $B=28{,}074$ textures gathered from public sources~\cite{Cimpoi2014Describing, Zhou2017Scene}, resize it to the camera resolution of $1920\times1080$, and sample between $5$ and $15$ cards to place on it.
For each card, we select a reference image at random, apply a random scale and in-plane rotation, and paste it at a random location, allowing partial overlaps so that the model sees crowded boards.
We automatically record the card annotation as its four corner coordinates with a single ``card'' class label.
We then apply randomized augmentations, each with a fixed probability, to narrow the gap to real footage: localized lighting effects that brighten parts of the scene and introduce saturated highlights, imitating spotlights, the striped pattern of neon tubes, and the specular glare that glossy cards produce ($p=0.33$), global saturation changes ($p=0.10$), additive and impulse noise ($p=0.20$), and a perspective warp that displaces the image corners within a margin ($p=0.20$).
The dataset comprises $18{,}000$ training, $1{,}000$ validation, and $1{,}000$ test generated images.

The orientation dataset is produced by the same compositing process restricted to a single card per image, rendered at the classifier's input resolution of $224\times224$.
Each card is placed close to upright, within $\pm10^\circ$, and half of the crops are then rotated by $180^\circ$; the applied rotation automatically determines the binary label.
This yields a balanced set of upright and inverted crops, $20{,}000$ in total, including a validation and test split.

The identification dataset provides, for each card in the reference set, several views of that card under varied appearance.
For every reference card we generate six $224\times224$ crops using the augmentations described above, so that the network sees each card under different lighting, color, and noise while its identity is fixed.
Of the six crops per card, five are used for training and one for validation, giving $5K$ training and $K$ validation crops, with $122{,}160$ images in total.%

\subsection{Real evaluation dataset}
\label{sec:realdata}

Synthetic data lets us train without manual annotation, but it cannot fully reproduce a real game: physical cards under real lighting, captured by consumer cameras, with the imperfections the recognition path must tolerate in deployment.
To measure how well the models transfer to these conditions, that is, the Sim2Real gap, we collect and manually annotate a dataset of real trading card scenes, including sleeved cards and holofoil prints.
This dataset is used only for evaluation; no model is trained or fine-tuned on it.
We capture real scenes of the Pok\'emon trading card game with a consumer webcam (Insta 360 Link 2C) mounted zenithally above the board, recording at a resolution of $1920\times1080$.
The scenes are captured under varied real lighting and deliberately include crowded boards, partially occluded cards, and face-down cards, so that the evaluation reflects realistic rather than idealized play.
In total, the dataset contains $51$ images with $832$ manually annotated card instances.
Each instance is annotated manually with an oriented bounding box, given as four corner coordinates, and the identity of the card, expressed using the same reference identifiers as the card database.
The annotations were produced by the authors with a custom annotation tool and cross-checked.

%% file: tables/datasets.tex
\begin{table}[t]
  \centering
  \caption{\textbf{Composition of our datasets.} The three synthetic datasets are generated automatically from the reference card images and annotated without human intervention; the real dataset is captured and manually annotated, and used only for evaluation. $K$ denotes the number of cards in the reference set ($K=20{,}360$ for Pok\'emon).}
  \label{tab:datasets}
  \resizebox{\textwidth}{!}{%
  \begin{tabular}{lllll}
    \toprule
    Dataset & Purpose & \#\,Images & Resolution & Annotation \\
    \midrule
    Synthetic detection    & train/val/test & $18{,}000$ / $1{,}000$ / $1{,}000$ & $1920\times1080$ & oriented box, \texttt{card} \\
    Synthetic orientation  & train/val/test      & $18{,}000$ / $1{,}000$ / $1{,}000$               & $224\times224$   & upright / $180^\circ$ \\
    Synthetic identification & train/test    & $5K$ / $K$                          & $224\times224$   & card identity \\
    \midrule
    Real            & evaluation     & $51$ ($832$ inst.)        & $1920\times1080$ & oriented box + identity \\
    \bottomrule
  \end{tabular}%
  }
\end{table}

%% file: sections/5_experiments.tex
\section{Experiments}
\label{sec:experiments}

We evaluate our TCG-AR pipeline along two axes: the performance of the recognition stages, and the runtime throughput of the pipeline, the latter determining whether the system is usable for live streaming.
All experiments are run on a single machine equipped with an AMD Ryzen~5~3600X CPU and an NVIDIA RTX~2080~Ti GPU, which corresponds to a consumer-grade streaming setup.
We report performance on both the synthetic test data and on the real evaluation dataset, which together quantify the Sim2Real gap, and we report throughput across the recognition stages and the output forms that drive it. To justify our architectural choices, we further compare five oriented detectors, four orientation backbones, and two identification heads, as reported in \cref{tab:recognition}. All details on model training parameters can be found in supplementary material.

\subsection{Detection and orientation}
\label{sec:exp-detection}

\input{tables/recognition}

Regarding detection, we compare five oriented detectors, all with a ResNet-50 backbone and trained under identical data and training parameters, so the comparison isolates the architecture. 
On real data, all five models reach essentially the same mAP of about $0.90$ (\cref{tab:recognition}), which confirms that card detection transfers robustly from synthetic to real regardless of the detector. The architectures separate mainly on the synthetic test set, where the two-stage Oriented R-CNN is clearly strongest ($0.91$~mAP, against $0.82$ for the one-stage RetinaNet and FCOS and lower for the heavier RoI Transformer and Gliding Vertex), and on speed, where FCOS is fastest on the GPU and the one-stage detectors are fastest on the CPU. We keep Oriented R-CNN in the deployed pipeline for its balance of accuracy and speed ($0.90$ real mAP at $14$~FPS on the GPU).
Regarding orientation, the $180^\circ$ classifier is a simpler binary task: all four backbones reach $\ge 98\%$ accuracy on real crops and essentially $100\%$ on synthetic (\cref{tab:recognition}). %

\subsection{Identification}
\label{sec:exp-identification}
\Cref{tab:recognition} shows that the choice of descriptor head is decisive on real data. 
The triplet head, although adequate on synthetic crops, collapses on real cards, reaching only $1.7\%$ top-1 against the full reference set, whereas the ArcFace head reaches $85.1\%$ under the same open-set protocol. 
The candidate-set size is the second lever. Without restriction, each crop is matched against the full reference set of $K=20{,}360$ cards; with the deck-list restriction, the candidate set shrinks to the $631$ unique cards. 
On the real dataset, this raises ArcFace top-1 accuracy from $85.1\%$ to $96.4\%$ by removing cards that cannot appear. 
Its effect on identification speed is small, however (\cref{tab:recognition}): the per-frame cost is dominated by the embedding network rather than the nearest-neighbor search, which is cheap even over the full reference set on the GPU.
Finally, the current models proved robust across the lighting conditions we captured.%

\subsection{Rendering and output}
\label{sec:exp-rendering}

\input{tables/throughput}

The rendering path runs concurrently with recognition.
\Cref{tab:throughput} reports its cost per output form on a single view. Raw output is essentially free; the augmented form composites the sprites at $118$~view-FPS, and the broadcast form, which additionally composes the side panels, runs at $34$~view-FPS; producing all three forms of a view together still sustains $25$~FPS.
Meanwhile the recognition path refreshes the game state at around $10$~FPS on the GPU.%

\subsection{Qualitative results}
\label{sec:exp-qualitative}

\begin{figure}[t]
  \centering
  \includegraphics[width=\linewidth]{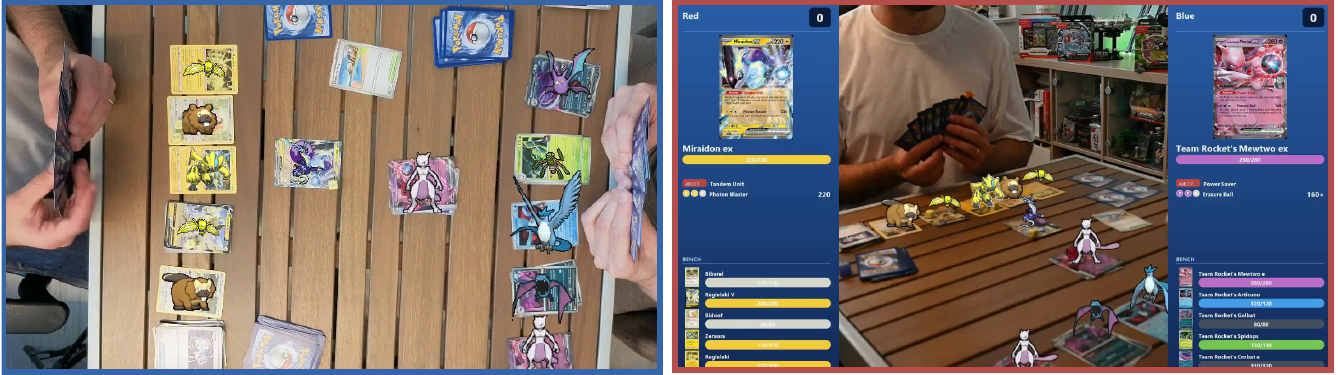}
  \caption{\textbf{Qualitative results.} Augmented output produced by TCG-AR. Left: the zenithal view after augmentation, with a virtual model rendered onto each recognized card. Right: Auxiliary view augmented from the same game state through registration with side panels showing card information. Several cards carry damage counters and status markers that occlude part of their design. TCG-AR identifies them correctly.%
  }
  \label{fig:qualitative}
\end{figure}

\Cref{fig:qualitative} shows the zenithal view before and after augmentation together with auxiliary views rendered from the same game state and the broadcast composition. 
Because cards are augmented during play rather than presented in isolation, they are frequently partially covered: damage counters, status markers, and neighboring cards placed on top may occlude parts of the artwork and text.
Our identification stage still retrieves the correct card in these cases, matching it from the visible remainder rather than from any single region of the card.
Furthermore, we observed stable behavior across the lighting conditions we captured, in line with the robustness of the recognition stages reported above; the residual errors are those inherited from recognition, namely occasional misses on heavily cluttered boards and confusions between visually near-identical cards.

\subsection{Discussion and limitations}
\label{sec:discussion}

Taken together, the results show that a markerless RGB-only pipeline can augment a trading card game in real time, but they also delineate where it currently falls short.
Detection now transfers well from synthetic to real data, and the ArcFace descriptor makes identification robust on real crops  (\cref{tab:recognition}); the deck-list restriction further improves identification when deck lists are available.
The remaining components are deliberately simple and limit the system at the edges: registration relies on a planar homography between fixed cameras and its automatic estimation is the least reliable stage in practice, while temporal matching compares only consecutive states and therefore loses cards under occlusion or a passing hand; these would benefit from more robust feature matching and a short temporal voting window, respectively.
Finally, the rendering path is restricted to two-dimensional sprites; extending it to three-dimensional models, and validating it on trading card games beyond the one studied here, are the natural next steps toward the accessible streaming tool this work aims to provide.

%% file: tables/recognition.tex
\begin{table}[t]
  \centering
  \caption{\textbf{Game state recognition quantitative results.} Five oriented detectors, four orientation backbones, and the two identification heads are evaluated on both the synthetic and real test sets, with reported per-stage speed (frames per second on a board of $14$ cards). 
  \emph{Real (deck)} restricts the identification candidate set to the $631$ unique cards present in the real data; identification speed is reported as full\,/\,deck.  The deployed pipeline uses Oriented R-CNN, EfficientNet-B0, and ArcFace.}
  \label{tab:recognition}
  \footnotesize
  \setlength{\tabcolsep}{6pt}
  \renewcommand{\arraystretch}{1.2}
  \resizebox{\textwidth}{!}{%
  \begin{tabular}{@{}l ccc cc@{}}
    \toprule
    & Synthetic & Real & Real (deck) & FPS$_{\text{GPU}}$ & FPS$_{\text{CPU}}$ \\
    \midrule
    \textit{Detection} &
    \multicolumn{2}{c}{ \textit{mAP\,/\,Recall}} & & \\
    Oriented R-CNN~\cite{Xie2021Oriented}      & $0.91$\,/\,$0.91$ & $0.90$\,/\,$0.90$ & -- & $14.0$ & $0.6$ \\
    Rotated RetinaNet~\cite{Lin2017Focal}      & $0.82$\,/\,$0.83$ & $0.91$\,/\,$0.92$ & -- & $14.1$ & $0.9$ \\
    Rotated FCOS~\cite{Tian2020FCOS}           & $0.82$\,/\,$0.81$ & $0.90$\,/\,$0.91$ & -- & $15.9$ & $0.9$ \\
    RoI Transformer~\cite{Ding2019Learning}    & $0.73$\,/\,$0.75$ & $0.91$\,/\,$0.90$ & -- & $13.1$ & $0.4$ \\
    Gliding Vertex~\cite{Xu2021Gliding}        & $0.64$\,/\,$0.68$ & $0.90$\,/\,$0.90$ & -- & $14.7$ & $0.4$ \\
    \midrule
    \textit{Orientation} &
    \multicolumn{2}{c}{ \textit{Accuracy}} & & \\
    EfficientNet-B0~\cite{Tan2019EfficientNet} & $100.0$ & $98.7$ & -- & $119$ & $2$ \\
    ResNet-18~\cite{He2016DeepResidual}        & $99.9$  & $98.9$ & -- & $144$ & $4$ \\
    MobileNetV3-S~\cite{Howard2019Searching}   & $100.0$ & $98.7$ & -- & $203$ & $11$ \\
    ShuffleNetV2~\cite{Ma2018ShuffleNet}       & $100.0$ & $98.1$ & -- & $108$ & $7$ \\
    \midrule
    \textit{Identification} &
    \multicolumn{3}{c}{\textit{Top-1 accuracy\,/\,Top-5 accuracy}} & \\
    Triplet~\cite{Schroff2015FaceNet}          & $48.5$\,/\,$71.7$ & $1.7$\,/\,$4.4$ & $17.3$\,/\,$36.8$ & $57$\,/\,$60$ & $1.3$\,/\,$1.3$ \\
    ArcFace~\cite{Deng2022ArcFace}             & $97.0$\,/\,$99.5$ & $85.1$\,/\,$96.2$ & $96.4$\,/\,$99.2$ & $53$\,/\,$55$ & $1.2$\,/\,$1.2$ \\
    \bottomrule
  \end{tabular}}
\end{table}

%% file: tables/throughput.tex
\begin{table}[t]
  \centering
  \caption{\textbf{Rendering throughput}, measured on the evaluation machine on one $1920\times1080$ view and a busy board of $14$ cards. \emph{Composite} is the cost of drawing the virtual characters, \emph{resize\,+\,copy} the cost of preparing the preview and the shared-memory frame, and \emph{total} their sum. 
  The broadcast form's side panels are cached, so a warm frame (cache hit) is much cheaper than a cold one (panel rebuild). 
  \emph{Streams@\,30\,FPS} is how many such views fit in the $40$\,ms display budget. 
  Each form is produced only when selected, so producing fewer forms frees both rendering and encoder budget.}
  \label{tab:throughput}
  \footnotesize
  \setlength{\tabcolsep}{10pt}
  \renewcommand{\arraystretch}{1.2}
  \resizebox{\textwidth}{!}{%
  \begin{tabular}{@{}lccccc@{}}
    \toprule
    Output form & Composite (ms) & Resize\,+\,copy (ms) & Total (ms) & View-FPS & Streams @\,30\,FPS \\
    \midrule
    Raw               & $0.0$  & $1.9$ & $1.9$  & $532$ & $17$ \\
    Augmented (AR)    & $6.6$  & $1.9$ & $8.5$  & $118$ & $3$ \\
    Broadcast (warm)  & $27.4$ & $1.9$ & $29.3$ & $34$  & $1$ \\
    Broadcast (cold)  & $35.9$ & $1.9$ & $37.7$ & $27$  & $<1$ \\
    \midrule
    All three forms   & $34.0$ & $5.6$ & $39.7$ & $25$  & $<1$ \\
    \bottomrule
  \end{tabular}}
\end{table}

%% file: sections/6_conclusion.tex
\section{Conclusion}
\label{sec:conclusion}

We presented \mbox{TCG-AR}, a real-time pipeline that augments trading card games from ordinary RGB cameras, without the chips, or instrumented surfaces that previous systems require.
From a single zenithal view, the pipeline detects, orients, and identifies the played cards to estimate the game state, and renders a virtual model onto each card across one or more registered views, and can compose a broadcast-style summary of the match, streaming the result to standard broadcasting software.
To train the recognition models without manual labeling, we introduced an automatic, game-agnostic procedure that generates annotated synthetic data from a reference set of card images, and we evaluated the trained models on a new manually annotated real dataset, measuring how they transfer to genuine play.
The design is organized around a real-time budget, separating a fast rendering path from a slower recognition path, so that the augmented video stays fluid while the game state is refreshed quickly enough for live viewing.
By relying only on commodity hardware and releasing our code, models, and datasets, we hope to bring augmented-reality streaming within reach of the broad community of trading card game players and streamers, and to provide a foundation for further work on vision-based augmentation of these games.

%% file: sections/7_appendix.tex
\section{Supplementary Material}
\label{sec:supplementary_material}

This appendix presents more information about the interface and card-rendering implementation, training details for the detection, orientation, and identification models, failure cases, and generalization.

\subsection{Interface and card rendering}

On launch, the interface lists the detected cameras, lets the operator designate the zenithal reference view, and runs the registration; when the automatic homography is unsatisfactory the operator can pick corresponding key points by hand to refine it. 
Each feed advertises its streaming address, so the augmented and broadcast outputs can be picked up directly by standard broadcasting software.

The sprite is drawn at an adjustable scale, set by default to twice the card size so that the model is clearly visible, and when several models overlap in a view they are composited back-to-front, ordered by their projected vertical position so that nearer models occlude farther ones. 
To convey the opposition between the two sides of the board, the models on one side are mirrored horizontally so that the two players' creatures face each other; since this is decided from the board position, the same orientation is applied consistently across all views. 
Cards that do not depict a creature, such as trainer and energy cards, are recognized and listed but left un-augmented, as they have no associated model.
The same entity may admit several visual variants, which the renderer exposes as selectable options, although such variants are specific to the game being augmented rather than to the method.
On the zenithal view the sprite is centred on its card, whereas on the auxiliary (side) views it is anchored by its base and rendered slightly smaller, so that the creature appears to stand on the card from an oblique viewpoint.

\begin{figure}[!ht]
  \centering
  \includegraphics[width=\linewidth]{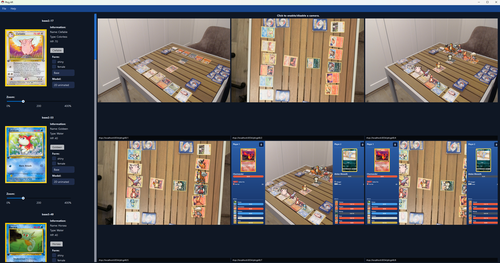}
  \caption{\textbf{GUI composition produced by \mbox{TCG-AR}.} The user can control which camera feeds to send to OBS. A side panel provides extra card annotation and control over the model size and forms, \eg, shiny or male versus female versions of the creature.}
  \label{fig:supp_side_panel}
\end{figure}

Beyond augmenting individual feeds, the rendering path can assemble a broadcast-style view that presents the game state as a television production would, illustrated in \cref{fig:qualitative}. 
A chosen feed is placed at the center, and a panel on each side summarizes one player's position: the active creature is shown with its card illustration, name, and game information, namely its health, type, and attacks with their energy costs, while the creatures on the bench appear below as a row of up to five cards. 
The cards in each panel are taken directly from the recognized game state, the active and benched cards being inferred from their positions on the board, whereas the information the system cannot observe, namely the players' names, scores, the venue, and each active creature's health and status, is entered by the operator through the interface described next. 
Because the panels depend only on the game state and not on the viewpoint, their rendered layer is computed once and shared across all views, and rebuilt only when the game state or the operator-entered information changes, which keeps the composition within the real-time budget. The composed view is streamed like any other output.
A recognized-cards side panel lists every detected card with its reference information and exposes the per-card rendering options (visual variant, scale, and whether a given card is augmented), illustrated in~\cref{fig:supp_side_panel}. 
A settings dialog collects machine-level parameters such as the streaming server path.

Furthermore, a small user interface lets the operator control the augmented output without editing code.
It displays the available feeds in a grid, showing for each camera its raw, augmented, and broadcast forms side by side, each independently selectable so that only the chosen feeds are produced and streamed, which saves computation.
A side panel lists the cards currently recognized, with their reference information, and exposes the per-card rendering options described above.
A separate panel collects the broadcast information the system cannot observe, namely the player names, scores, venue, and each active creature's current health and status, which feeds the broadcast composition above, and the interface also exposes minor conveniences such as displaying the streaming address of each feed.
This interface is intended to keep the system usable by streamers without technical expertise, in line with our goal of accessibility.

\begin{figure}[!ht]
  \centering
  \includegraphics[width=\linewidth]{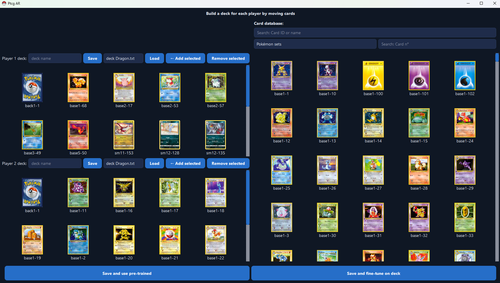}
  \caption{\textbf{Operator interface (deck-selection panel).} The full card database is shown as a searchable grid (right) from which a deck is assembled for each player (left); a card may belong to both decks, and identification is restricted to their union. 
  Decks can be named, saved and reloaded, and the operator chooses between using the pre-trained model as-is or fine-tuning it on the deck. 
  Shown for the Pok\'emon trading card game.}
  \label{fig:deck}
\end{figure}

Beyond the broadcast controls described above, the interface exposes the rest of the pipeline so that a streamer can operate it without editing code; \cref{fig:deck} shows the deck-selection panel. A setup panel presents the full card database as a searchable grid (by name, by number, and by set) from which the operator assembles a deck for each of the two players; a card may be placed in both decks, and identification is then restricted to the union of the two decks. 
Decks can be named, saved and reloaded across sessions, and the operator chooses whether to use the pre-trained model as-is on the restricted candidate set or to fine-tune it on the deck before starting.

\subsection{Implementation and training details}
\label{sec:supp-training}

This section reports the full training and evaluation setup for every model (detection, orientation, identification), so that all results in the main paper can be reproduced. 
Unless stated otherwise, all models are trained only on the synthetic training split and evaluated on the held-out synthetic test split and on the real set; no real data is used for training. 
After removing wrong scans (\cref{sec:supp-cleaning}) and adding the card back as one extra identity, the reference set contains $K\!=\!20{,}360$ cards.

\subsubsection{Hardware and software}
\label{sec:supp-env}
All experiments run on a single machine: AMD Ryzen~5~3600X CPU and one NVIDIA RTX~2080~Ti GPU. Software: Python~3.11, PyTorch~2.0.1 / torchvision~0.15.2 (CUDA~11.8); the detector uses mmrotate~0.3.4, mmdet~2.28.2 and mmcv-full~1.7.2. 
Image pre-processing uses OpenCV and Pillow. 
All backbones are initialised from ImageNet weights. 

\subsubsection{Card detection}
\label{sec:supp-detection}
The deployed detector is Oriented R-CNN~\cite{Xie2021Oriented} with a ResNet-50~\cite{He2016DeepResidual} backbone (ImageNet-pretrained, first stage frozen, batch-norm statistics frozen) and a feature pyramid~\cite{Lin2017Feature}, a single object class (\texttt{card}), and the \texttt{le90} angle convention. The four comparison detectors
(Rotated RetinaNet~\cite{Lin2017Focal}, Rotated FCOS~\cite{Tian2020FCOS}, RoI Transformer~\cite{Ding2019Learning}, Gliding Vertex~\cite{Xu2021Gliding}) reuse the \emph{exact} data pipeline, optimiser and schedule below; only the detector architecture changes (each is the stock mmrotate \texttt{r50\_fpn\_1x\_dota\_le90} model with \texttt{num\_classes} set to $1$). 
Images are resized to $1024\times1024$ (\texttt{RResize}); random rotated flips are applied horizontally, vertically and diagonally (each with probability $0.25$); pixels are normalised with
mean $(123.675,116.28,103.53)$ and std $(58.395,57.12,57.375)$ (RGB), then padded to a multiple of $32$.
SGD with learning rate $5\times10^{-3}$, momentum $0.9$, weight decay $10^{-4}$, gradient-norm clipping at $35$. 
A linear warm-up over $500$ iterations (ratio $1/3$) precedes a step schedule that divides the learning rate by $10$ at epochs $2$ and $4$. 
Training runs for $5$ epochs with a batch size of $2$ images on one GPU ($\sim\!9{,}000$ iterations/epoch over the $18{,}000$ training images); a checkpoint is saved every epoch. 
The RPN uses anchors of scale $8$ with aspect ratios $\{0.5,1,2\}$ over strides $\{4,8,16,32,64\}$; the RoI head uses rotated RoI-align (output $7\times7$) and two shared fully connected layers of width $1024$. 
At test time boxes are kept above a score of $0.8$ with rotated NMS at IoU $0.5$.

\subsubsection{Card orientation}
\label{sec:supp-orientation}
The orientation classifier is a binary (upright vs.\ $180^\circ$) network on $224\times224$ crops. The deployed model is EfficientNet-B0~\cite{Tan2019EfficientNet}; the comparison backbones are ResNet-18~\cite{He2016DeepResidual}, MobileNetV3-Small~\cite{Howard2019Searching} and ShuffleNetV2~\cite{Ma2018ShuffleNet}.
Each is ImageNet-pretrained with its classification head replaced by a two-way linear layer; everything else is identical so the comparison isolates the backbone.
Resize to $224\times224$, random horizontal flip ($p=0.5$), the synthetic augmentation suite of \cref{sec:synthetic} (lighting, saturation, noise, perspective), random sharpness adjustment ($p=0.5$), then \texttt{ToTensor} and ImageNet normalisation (mean $(0.485,0.456,0.406)$, std $(0.229,0.224,0.225)$). 
Validation/test use only resize and normalization.
Adam with learning rate $10^{-3}$, a step scheduler (decay every $5$ epochs), cross-entropy loss, batch size $16$, for $10$ epochs; a checkpoint is saved every epoch.

\subsubsection{Card identification}
\label{sec:supp-identification}
Both identification heads share a ResNet-50 backbone (ImageNet-pretrained) on $224\times224$ inputs and are trained on the synthetic identification split (six augmented crops per card: five for training, one held out for validation). The train augmentation is: resize to $224\times224$, Gaussian blur (kernel $5\!-\!9$, $\sigma\in[0.1,5]$), random sharpness adjustment ($p=0.5$), \texttt{ToTensor}, ImageNet normalisation.
At test time a card is identified by nearest neighbour to a set of reference descriptors, one per card, each computed once from the card's reference image.

For triplet, the backbone's final layer is replaced by $\mathrm{Linear}(2048,512)\!\rightarrow\!\mathrm{ReLU}\!\rightarrow\!\mathrm{Linear}(512,128)$, giving a $128$-d descriptor. Training uses the triplet margin loss (margin $1.0$, Euclidean distance) with hard-negative mining (the negative for each anchor is a visually close card rather than a random one), Adam at learning rate $10^{-3}$ with a step schedule (factor $1/3$ every $4$ epochs), batch size $16$, for $20$ epochs. Retrieval uses Euclidean distance.

For ArcFace, the final layer is replaced by $\mathrm{BN}(2048)\!\rightarrow\!\mathrm{Dropout}(0.5)\!\rightarrow\!\mathrm{Linear}(2048,512)\!\rightarrow\!\mathrm{BN}(512)$, giving a $512$-d embedding. Training uses an additive angular-margin softmax (ArcFace~\cite{Deng2022ArcFace}) with scale $s=30$ and margin $m=0.50$, one output class per card. 
Optimisation is SGD with learning rate $10^{-3}$, momentum $0.9$, weight decay $5\times10^{-4}$, and a cosine-annealing schedule (over the $20$ epochs, minimum learning rate $10^{-3}\!\times\!10^{-3}$); cross-entropy on the margin logits; batch size $16$. A checkpoint is saved every epoch. 
We found that the Adam/step recipe of the triplet head collapses the learning rate before the $\sim\!20$k-class margin head converges; the SGD/cosine recipe above is required. 
At test time the ArcFace classifier head is discarded and the $512$-d embedding is compared by cosine similarity.

\subsubsection{Hard negative sampling for identification}
\label{sec:supp-hardneg}
\begin{figure}[t]
  \centering
  \includegraphics[width=\linewidth]{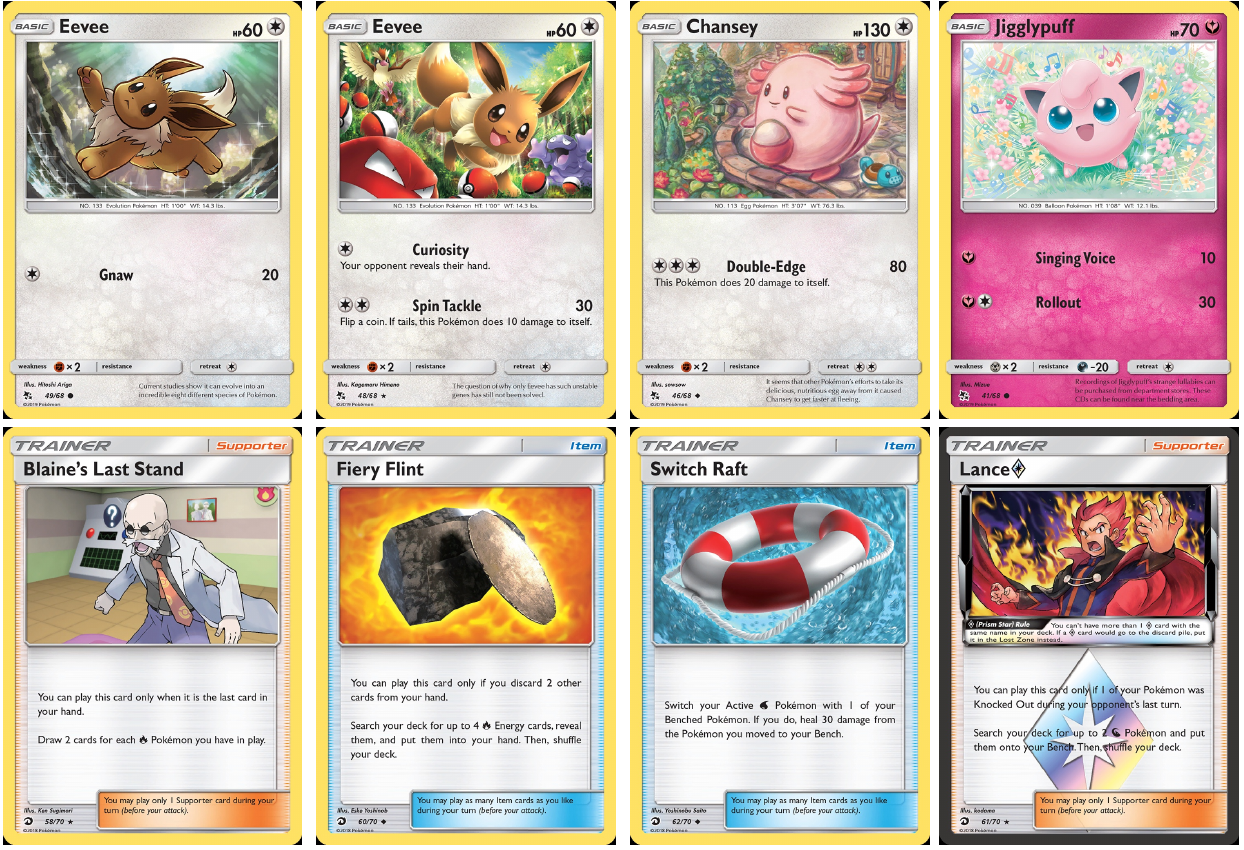}
  \caption{\textbf{Hard-negative tiers used for the triplet head.} For a creature anchor (left), negatives are drawn with decreasing probability from the same species (a different print of the same creature, \ie the hardest case), the same energy type, and then any creature. 
  Trainer and Energy anchors draw from their own category. The curriculum forces the descriptor to separate look-alike cards rather than easy random pairs. 
  Shown for the Pok\'emon trading card game.}
  \label{fig:hard-negatives}
\end{figure}

The triplet head's difficulty comes almost entirely from cards that look alike: different prints of the same creature, cards that share an artwork frame, or the many near-identical Trainer and Energy cards.
Drawing the negative of each anchor uniformly at random would therefore spend most of the training budget on trivially easy pairs. 
Instead, we sample negatives with a category-aware curriculum driven by the card metadata, so that
the loss concentrates on the confusions that actually occur. 
For a creature (Pokémon) anchor, with probability $\approx\!0.5$ the negative is another card of the \emph{same
species}, \ie, same national Pokédex number(s), type and subtype, a different print or illustration of the same creature, the hardest case (and we raise this probability when the anchor has no Pokédex number, such as a special art card); with probability $\approx\!0.2$ it is a card of the \emph{same energy type} but a different species; with probability $\approx\!0.1$ it is any other creature; and otherwise the negative is drawn uniformly. 
Trainer and Energy anchors, which form large visually homogeneous families, draw their negative from their own category ($80\%$ of the time) before falling back to a  random card. \Cref{fig:hard-negatives} shows an anchor with one negative sampled from each tier. 
This sampling is used for the triplet head; the ArcFace head obtains an analogous effect implicitly, since its angular-margin softmax pushes every card away from all $\sim\!20$k class centers.

Finally, when deck lists are available, the descriptor network can additionally be fine-tuned for $10$ epochs at learning rate $10^{-4}$ (triplet batch size $8$; ArcFace resumed from the base weights) on the restricted candidate set before a session, further separating the cards that will actually be encountered. 
This is optional and is \emph{not} used for the main-paper numbers, which evaluate the base models.
Furthermore, it is interesting to note that our model generalizes well to cards written in other languages than English, such as French or German, who share similar illustrations.

\subsubsection{Reference-set cleaning}
\label{sec:supp-cleaning}
A small fraction of reference images are wrong scans whose stored picture is the card back. 
We embed every reference image with the base ArcFace model and flag those whose cosine similarity to the embedding of a known card back exceeds $0.85$; flagged cards are excluded from training, retrieval and evaluation. The card back itself is retained as a separate identity so that face-down cards are recognized as such.

\subsubsection{Evaluation protocols}
\label{sec:supp-eval}

\noindent\textit{Detection.} We report DOTA-style mean average precision (AP at IoU $0.5$) and recall, on the synthetic test split ($1{,}000$ images) and the real set ($51$ images, $832$ annotated instances); for the real set the test data paths are overridden to the real annotations, leaving everything else identical. 
Per-stage frame rate is measured by timing single-image inference over $100$ images on the GPU and the CPU.

\noindent\textit{Orientation.} We report top-1 accuracy of the binary upright/$180^\circ$ decision on the synthetic test split ($1{,}000$ crops) and on $636$ crops derived from the real annotations (each ground-truth-upright real crop kept straight or rotated $180^\circ$, balanced). Forward frame rate is measured on the GPU and the CPU.

\noindent\textit{Identification.} We report top-1 and top-5 accuracy by nearest-neighbour retrieval against the reference descriptors (cosine for ArcFace, Euclidean for triplet). 
Identification is evaluated on \emph{ground-truth-upright} crops (orientation taken from the annotation), so that it is measured independently of the orientation model. 
On the synthetic test data each card's held-out sixth crop is queried against the full reference set. On the real set ($636$ annotated crops) we report two regimes: the open set (all $20{,}360$ cards as candidates) and the deck-restricted set (candidates limited to the $631$ unique cards present in the real data). 

\noindent\textit{Throughput.} Per-stage recognition speed is reported as full zenithal frames per second assuming a busy board of $14$ cards, on the GPU and the CPU; rendering cost is reported per output form (raw, augmented, broadcast).

\subsection{Failure cases}
\label{sec:supp-failure}
The two recognition stages fail in different ways. The detectors behave as detectors of rectangular card-like regions: they are most reliable on well-separated, face-up cards and degrade on heavily cluttered arrangements, where they both miss cards and produce false positives, and where overlapping cards occasionally merge into a single box. 
Face-down
cards are detected but, having no face, are identified as the card back. 

\begin{figure}[!ht]
  \centering
  \includegraphics[width=\linewidth]{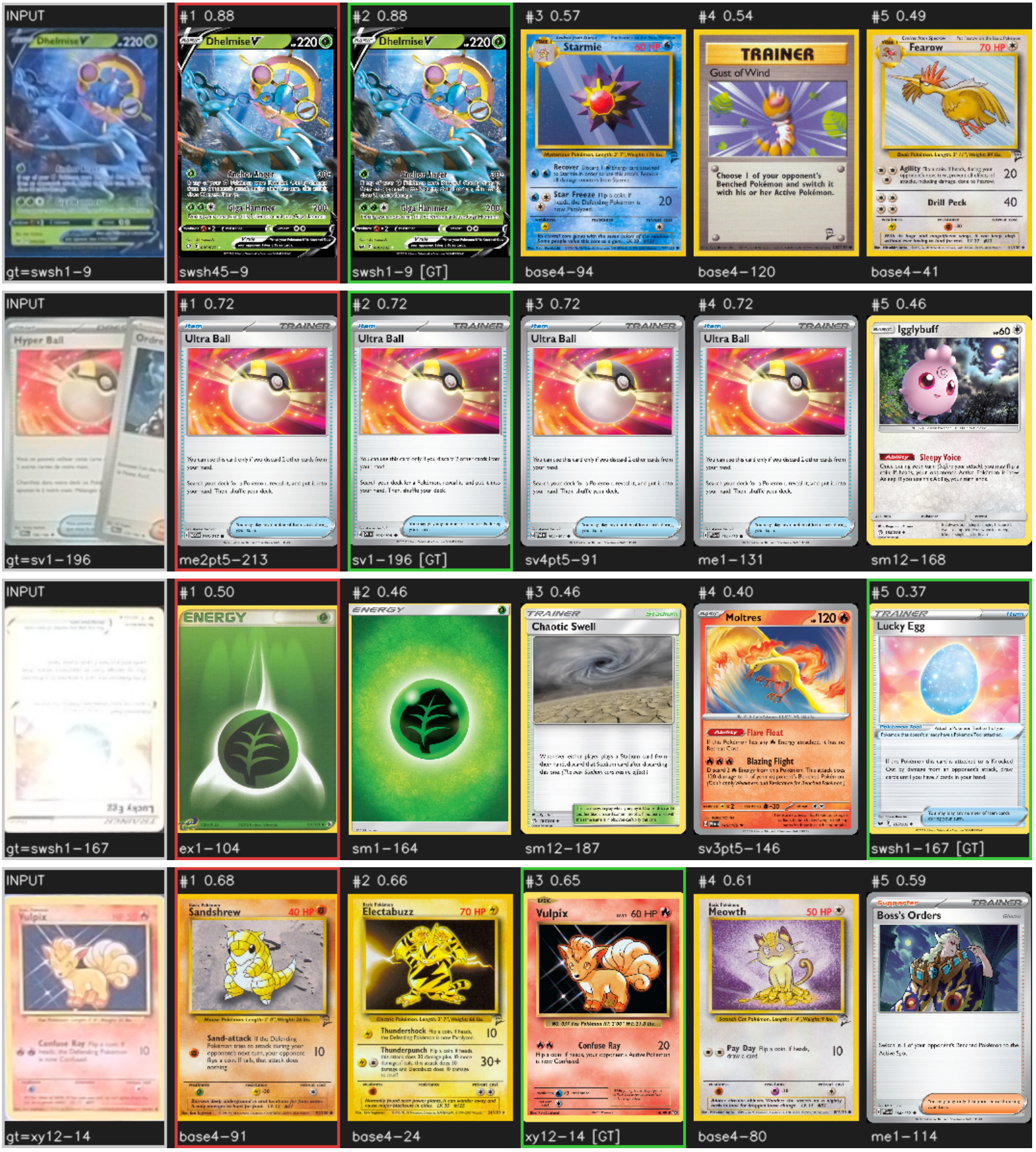}
  \caption{\textbf{Representative identification errors (real set, ArcFace).} Each row shows a query crop (left) and the top-five retrieved reference cards; the ground-truth card, when present, is outlined. Most errors are confusions between near-identical cards, \eg, different prints of the same creature or cards sharing an illustration frame, so the correct card usually remains within the top five.}
  \label{fig:id-errors}
\end{figure}

The errors of the deployed ArcFace head are dominated by genuinely look-alike cards rather than by obvious mistakes. On the $636$ real crops, the open set produces $95$ top-1 errors, of which $71$ ($75\%$) still place the correct card within the top five retrieved references; only $24$ are top-5 misses. Restricting the candidate set to the cards present in the decks removes most of these confusions, leaving $23$ top-1 errors and only $5$ top-5
misses. \Cref{fig:id-errors} shows representative cases: the query is almost always retrieved alongside other prints of the same creature or cards sharing the same frame, and the few top-5 misses correspond to crops degraded by motion blur, glare or strong perspective rather than to a confusion between unrelated cards. This is the behaviour the hard-negative curriculum of \cref{sec:supp-hardneg} targets, and it explains why a small candidate restriction (the deck list) recovers most of the lost accuracy.

The registration is the least robust component.
It is a simple planar mapping between fixed views performed at the beginning of the capture sequence. It is not a full six-degree-of-freedom camera pose estimation: it assumes the board is planar and the cameras do not move after initialization, which holds for a typical tabletop streaming setup but would not accommodate freely moving cameras.
The automatic estimation is also the least reliable component of the pipeline in practice, mostly due to not enough distinctive keypoints in typical card game scenarios at the beginning of the game.
The fallback manual annotation proves to be fast, with around 30 seconds required to annotate at least 4 common keypoints across views.
In practice, since the cards themselves are richly textured, placing a few of them on the bare table before initialization gives enough distinctive keypoints to register the views.
A more principled alternative, which we leave to future work, is to register directly from the cards rather than the background: each detected card supplies four coplanar corners, and since all cards lie on the table plane, corner correspondences between views determine the same planar homography without relying on background texture at all.

\subsection{Assessing generalization}
\label{sec:supp-generalization}
A practical concern for a deployed system is whether it keeps working as new cards are released, since the trading card game grows continuously. 
Identification is designed for this: the ArcFace classifier head, which has one output per training card, is discarded at test time, and recognition is performed by nearest-neighbour retrieval against a per-card reference descriptor computed directly from the card's official image. 
A card the network never saw during training can therefore be recognised as soon as its reference image is added to the gallery, with no retraining. 
This is the property that lets the same models cover the full $20{,}360$-card reference set while being trained on synthetic crops alone. 
A more direct protocol freezes the embedding trained on cards released up to a cutoff date and evaluates retrieval on the ${>}2{,}000$ cards released afterward, which the network has never seen; retrieval remains accurate on these unseen cards.

\subsection{Scope of the recognized game state}
\label{sec:supp-scope}
The recognized game state is the set of visible, face-up cards, each with a position, an in-plane orientation, and an identity.
A physical trading card game board carries more information than this, and we make the boundary explicit here.
Several mechanics stack cards so that only the top one is visible: energy cards are placed beneath a Pokémon to power its attacks, and an evolution keeps its lower stages under the current one for the rest of the game.
A zenithal camera sees only the top card, so attachments and the hidden stages of an evolution chain cannot be recovered from a single frame, and the pipeline treats what is beneath the top card as unobserved.
Damage counters, status markers, and dice likewise convey numeric and categorical state that is not part of a card's identity.
The pipeline is robust to these as occluders, identifying a card from its visible remainder even when they cover part of it (Sec.~\ref{sec:exp-qualitative}), but it does not read their values; a creature's remaining health or a poison or paralysis condition is therefore entered by the operator, and in-plane rotation is the only status signal read directly from the board.

Finally, tracking associates cards between consecutive game states by nearest-position matching, which stabilizes identities frame to frame but keeps no longer history.
A card that becomes hidden, whether covered by a hand, another card, or a marker, leaves the state until it is visible again, at which point it is re-detected with no memory of its earlier identity.
Persisting identities across occlusion, and reconstructing attachment and evolution relationships over the course of a game, would require a temporal model of the board and are left to future work.
Energy attachment is a concrete case where these limits combine.
A player attaches an energy from hand in a quick motion, often under a second, and tucks it under the Pokémon so that only a strip remains visible, with no standard placement across players.
The recognition path refreshes at roughly four frames per second, so the placement and the hand passing over it can fall entirely between two recognition frames.
Even once the board settles, the energy is mostly occluded and in no canonical orientation, so the detector and identifier may not recover it at all.
Finally, an attachment is a relationship between two cards, ``this energy powers that Pokémon,'' whereas the recognized state is a flat set of cards with a position, orientation, and identity, and has no field to express it.
The system therefore does not capture energy attachments: at best it reports the visible strip as a separate, possibly unidentified card, and more often the energy is absent from the state entirely.
Representing such relationships, and sampling the board densely enough to catch fast placements, would require both a temporal model and a relational state, and are left to future work.

\subsection{Rendering assets and licensing}
\label{sec:supp-assets}
The rendering path pairs each recognized card identity with a renderable model drawn from an asset database.
This database is not part of our system and is assembled at the user's end from external sources; we redistribute none of its contents.

For the Pokémon demonstration, the database combines two kinds of external data.
Card metadata and reference images, used both for identification and for the card panels of the broadcast view, are obtained through the Pokémon TCG API~\cite{Backes2021PokemonTCGSDK} and its associated card database.\footnote{\url{https://dev.pokemontcg.io}}
The creature models composited onto the cards are two-dimensional animated sprites collected from a community sprite repository,\footnote{\url{https://projectpokemon.org/home/docs/spriteindex_148}} indexed against the National Pok\'edex ordering.\footnote{\url{https://www.pokepedia.fr/Liste_des_Pok\%C3\%A9mon_dans_l\%27ordre_du_Pok\%C3\%A9dex_National}}
Both are fetched by the user at setup time; the system stores only the mapping from card identity to asset, not the assets themselves.

The creatures, their sprites, and the card artwork are the intellectual property of the game's publisher and are not openly licensed.
Our release therefore comprises the pipeline, the trained detection, orientation, and identification models, and the synthetic-data generator, but no copyrighted card images or creature assets; each user retrieves those from the sources above under the terms that apply to them.
The recognition models are trained on synthetically composited crops of the reference card images and do not embed redistributable copies of them, so the released models carry no card artwork.
We note that the community sprite sources are themselves of unclear redistribution status, which is a further reason we treat them strictly as user-supplied inputs rather than shipping them; a deployment that requires clean licensing could substitute any asset set keyed by card identity, as the renderer is agnostic to the origin of the models.

Because the asset database is external and could be incomplete, the renderer defines an explicit fallback for cards it cannot illustrate.
A card whose identity is recognized but for which the database holds no model, either a non-creature card such as a Trainer or Energy card or a creature absent from the current asset set, is detected, identified, and listed in the recognized-cards panel, but composited without an overlay.
Augmentation of the remaining cards on the board is unaffected, so a missing asset degrades coverage gracefully rather than interrupting the stream.